# Testing SPARUS II AUV, an open platform for industrial, scientific and academic applications

Marc Carreras, Carles Candela, David Ribas, Narcís Palomeras, Lluís Magí, Angelos Mallios, Eduard Vidal, Èric Vidal and Pere Ridao

Computer Vision and Robotics Institute, Universitat de Girona
Parc Científic i Tecnològic UdG, 17003, Girona, Spain.
marc.carreras@udg.edu; http://cirs.udg.edu

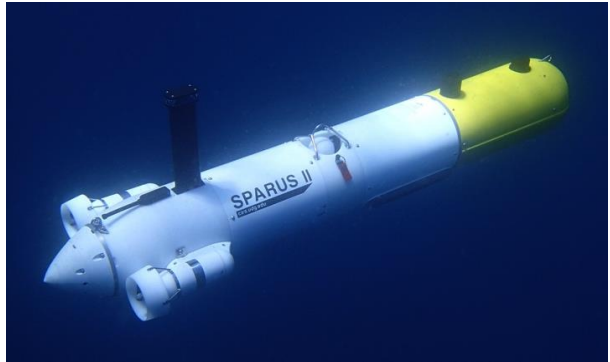

Figure 1. Sparus II AUV at sea.

*Abstract* – *This paper describes the experience of preparing and testing the SPARUS II AUV in different applications. The AUV was designed as a lightweight vehicle combining the classical torpedo-shape features with the hovering capability. The robot has a payload area to allow the integration of different equipment depending on the application. The software architecture is based on ROS, an open framework that allows an easy integration of many devices and systems. Its flexibility, easy operation and openness makes the SPARUS II AUV a multipurpose platform that can adapt to industrial, scientific and academic applications. Five units were developed in 2014, and different teams used and adapted the platform for different applications. The paper describes some of the experiences in preparing and testing this open platform to different applications.*

## I. INTRODUCTION

Commercial AUVs are mainly conceived to surveying applications in which large areas must be covered and the vehicle follows safe paths at safe altitudes. However, new advances in sonar technology, image processing, mapping and robotics will allow more complex missions, in which the AUV will be able to navigate at a closer distance from the seabed, it will react to the 3D shape of the environment, and it will even perform some autonomous intervention tasks. In this context, the Underwater Robotics Research Centre of the University of Girona has been developing several AUV prototypes during more than 15 years to achieve these new capabilities. The SPARUS Autonomous Underwater Vehicle (AUV) was conceived in the Underwater Robotics Research Centre (CIRS) of the University of Girona (Spain). The first version was designed in 2010 to participate in the European Student AUV competition, organized by CMRE in La Spezia (Italy). The robot won the competition and, since then, it has collaborated in several research projects. In 2013, a new version of the robot, SPARUS II AUV (Figure 1), was designed and 5 units were developed in 2014 for several research institutions. Three of them were specially developed for the euRathlon competition, in which teams learned the operation of the robot and adapted its use to the competition. This paper describes the experiences after more than one year testing the platforms and adapting them to different applications.

## II. SPARUS II DESIGN & EXPERIMENTATION

SPARUS II AUV (see Figures 2 and 3) is a lightweight hovering vehicle with mission-specific payload area and efficient hydrodynamics for long autonomy in shallow water (200 meters). It combines torpedo-shape performance with hovering capability. It is easy to deploy and to operate. The payload area can be customized by the end-user and it uses an open software architecture, based on ROS, for mission programming. Its flexibility, easy operation and openness makes the SPARUS II AUV a multipurpose platform that can adapt to industrial, scientific and academic applications. The key points of the vehicle are: a) torpedo-shape movement with efficient hydrodynamics and long autonomy; b) hovering capability for high maneuverability; c) lightweight vehicle, similar weight and size than underwater gliders; d) easy operation, which can be operated by 2 persons from any small boat; e) mission specific payload: open hardware for equipment integration; f) software architecture based on ROS: open software available for download. The AUV has three thrusters (two horizontal and one vertical) and can be actuated in surge, heave and yaw degrees of freedom (DOF). The vehicle is equipped with a navigation sensor suite including a pressure sensor, a doppler velocity log (DVL), an inertial measurement unit (IMU) and a GPS to receive fixes while at surface.

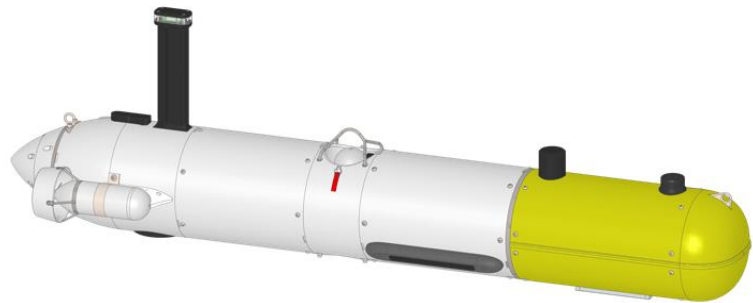

Figure 2. CAD design of Sparus II AUV.

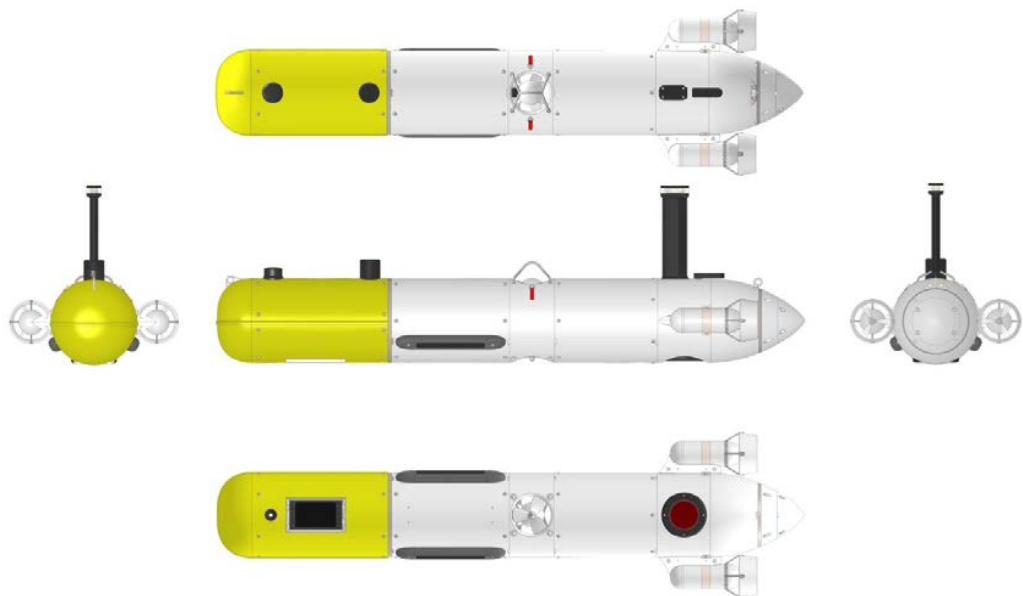

Figure 3. Different views of the research platform.

Application specific equipment is installed in the payload area, having also space inside the main hull for internal electronics. Regulated 12V (max power 40W) and 24V (max power 100W) and Ethernet and RS232 serial communication is available for new equipment. Connections are done using underwater connectors (Subconn compatible), which can be used for any required need. The payload can have up to eight connectors. The payload area can integrate any equipment having a maximum volume of 8 liters and a maximum weight of 7 kg. The non-required volume or weight is filled by foam and lead to maintain a constant total volume and weight, which is approximately 52 liters and 52 kg in air. Figure 4 shows an example of a real sonar and vision payload with the following equipment: Imagenex Delta T multibeam profiler sonar, down-looking HD FireWire Video Camera, 5 Imagenex echo-sounders for obstacle avoidance, Imagenex Side Scan Sonar, Tritech Micron Imaging Sonar and Evologics acoustic modem with USBL. Figure 5 shows the data obtained with the side scan sonar in a real trajectory.

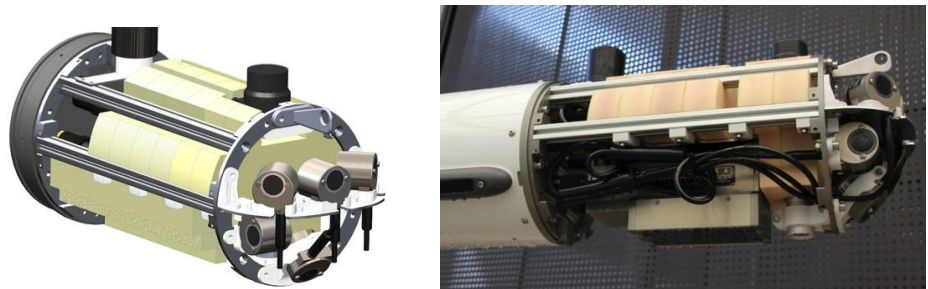

Figure 4. Design and picture of a complete sonar and video payload.

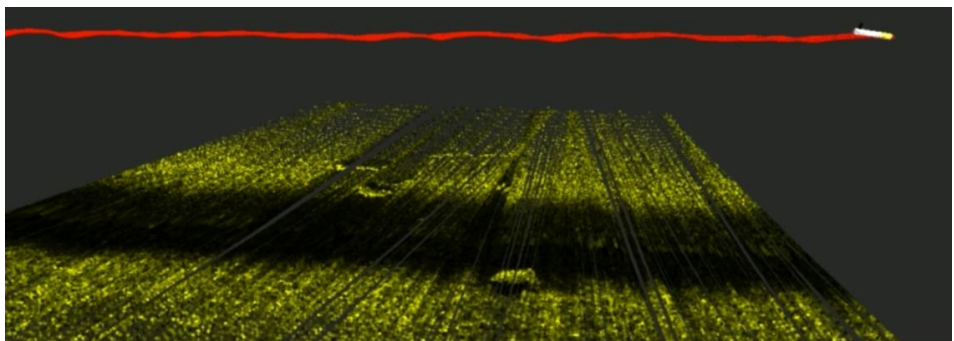

Figure 5. Example of side scan sonar data plotted according the navigation of the Sparus II AUV.

The same payload was used in euRathlon 2014 competition, see Figure 6, in which the University of Girona team won 4 out of the 5 proposed tasks. One of the task consisted on mapping an area of a harbour while detecting some landmarks by processing the images of a down-looking camera. Landmarks were some buoys that were placed in unknown locations and at different depths. Figure 7 shows the lawnmower trajectory that was pre-programmed and the position of the landmarks that were discovered during the movement. Also, it shows a bathymetry of the area that was generated using the multibeam profiler.

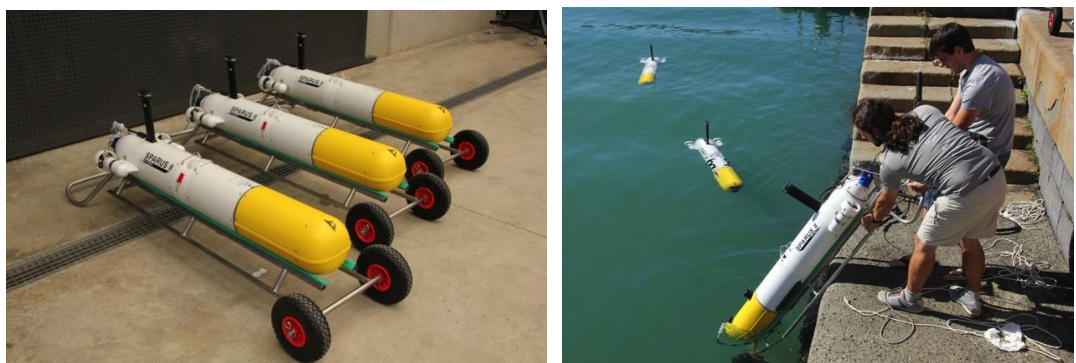

Figure 6. Sparus II AUV at euRathlon competition in 2014. Three vehicles were developed for teams participating in the competition. Deployment and recovery can be done easily.

In previous results, the AUV followed predefined trajectories as classical survey AUVs. Research in the University of Girona has also been conducted for developing the capabilities of future Inspection AUVs. In this sense, real-time mapping and motion planning was used for detecting unknown obstacles and moving through them. Experiments were done in the harbour of Sant Feliu de Guixols, in the external and open area of the harbour, in a breakwater structure that is composed of a series of concrete blocks of 14.5m long and 12m width, separated by a 4m gap with an average depth of 7m. The AUV was requested to achieve a way-point which was on the other side of the breakwater blocks and, without any previous knowledge, it was able to autonomously build the map of the blocks and to achieve the position. The hovering capability allowed a precise manoeuvring to face and pass through the corridor between the blocks. Figure 8 shows an image of the environment and the internal map that was build using the acoustic ranges from the multibeam, the 5 echo-sounders and imaging sonar. The trajectory followed by the AUV was at less than 2 meters of the concrete blocks, allowing a close inspection of the structure.

These results point out the suitability of using Sparus II AUV as an inspection AUV for industrials or scientific purposes.

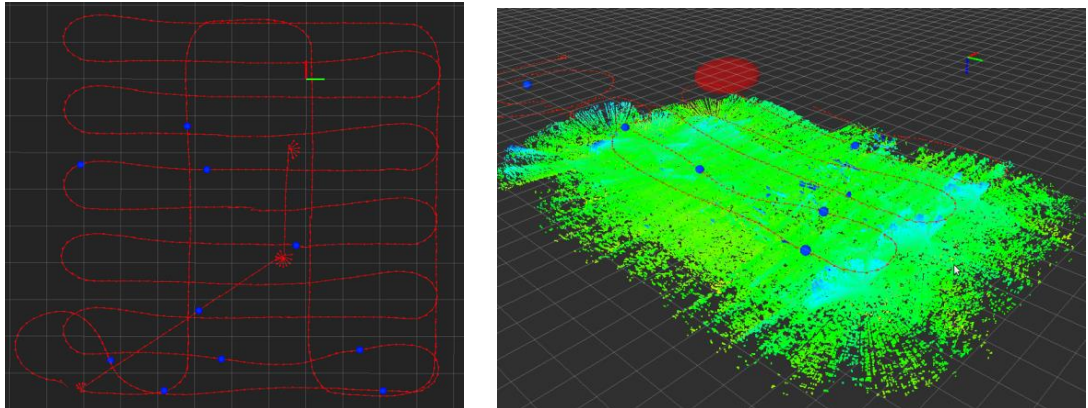

Figure 7. Lawnmower trajectory (left) and bathymetry (right) generated during euRathlon competition in 2014. Some blue dots represent the positions were landmarks were discovered.

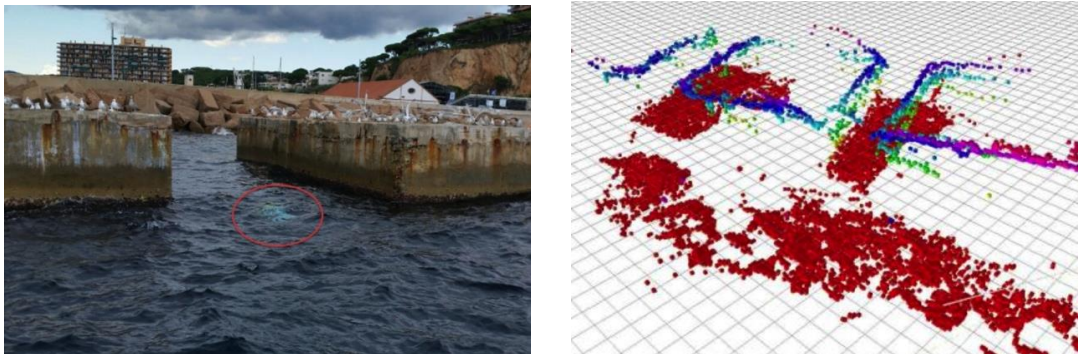

Figure 8. Sparus II AUV passing trough some breakwater blocks in an autonomous inspection task (left). The AUV is able to generate in real-time a 3D map of the environment and plan trajectories through it (right).

## III. CONCLUSIONS

This paper has presented the SPARUS II AUV as an open platform for integrating equipment and software for multipurpose applications. Five platforms were developed in 2014 and tested by 5 different teams with different payloads. The University of Girona was in charge of developing and teaching the teams the operation of the vehicle and the integration of new hardware and software. The experience was positive and the teams were able to integrate their own equipment and to operate the AUV successfully. The vehicle has also been used in different research project of the laboratory with successful results.

## IV. ACKNOWLEDGEMENTS